\documentclass[journal]{IEEEtran}
\IEEEoverridecommandlockouts    

\usepackage{graphicx}
\usepackage[font=footnotesize,labelfont=bf,skip=0pt]{caption}
\usepackage{epstopdf}
\usepackage{amssymb}
\usepackage{amsmath} 
\usepackage{enumerate}
\usepackage{etaremune}
\usepackage{setspace}
\usepackage{comment}
\usepackage{color}
\usepackage[table]{xcolor}
\usepackage{soul,xcolor}
\setstcolor{red}
\usepackage[normalem]{ulem}
\usepackage{wrapfig}
\usepackage[algo2e]{algorithm2e} 
\usepackage{algorithm} 
\usepackage{algpseudocode} 
\usepackage[hidelinks]{hyperref}
\usepackage{cite}
\definecolor{red}{RGB}{150, 11, 23}
\usepackage{soul}
\usepackage{booktabs}
\usepackage{float}
\usepackage{amsmath}
\usepackage{subfigure}
\usepackage{pgfplots} 
\pgfplotsset{compat=newest} 
\pgfplotsset{plot coordinates/math parser=false} 
\newlength\figureheight 
\newlength\figurewidth 
\usetikzlibrary{mindmap}
\usepackage{placeins}
\usetikzlibrary{arrows}
\usepgfplotslibrary{patchplots}

\definecolor{UBred}{RGB}{150, 11, 23}

 \usepackage{soul}
 \usepackage[switch]{lineno}

\usepackage{tcolorbox}
\newtcolorbox{mybox}{colback=yellow!50!white,colframe=yellow!100!black}
\usepackage{etoolbox}

\begin{document}

\title{\LARGE{{\bf KG-Planner: Knowledge-Informed Graph Neural Planning for Collaborative Manipulators}}}

\author{Wansong Liu$^{1}$, Kareem Eltouny$^{2}$, Sibo Tian$^{3}$, Xiao Liang$^{4}$, Minghui Zheng$^{3}$
    \thanks{This work was supported by the USA National Science Foundation  (Grants: 2026533/2422826 and 2132923/2422640). This work involved human subjects or animals in its research. The authors confirm that all human/animal subject research procedures and protocols are exempt from the University at Buffalo's review board approval.}
	\thanks{$^{1}$ Wansong Liu is with the Mechanical and Aerospace Engineering Department, University at Buffalo, Buffalo, NY 14260, USA. {\tt\small Email: wansongl@buffalo.edu}.}%
	\thanks{$^{2}$ Kareem Eltouny is with the Civil, Structural and Environmental Engineering Department, University at Buffalo, Buffalo, NY 14260, USA. {\tt\small Email: keltouny@buffalo.edu}.}
 	\thanks{$^{3}$ Sibo Tian and Minghui Zheng are with J. Mike Walker '66 Department of Mechanical Engineering, Texas A\&M University, College Station, TX 77843, USA. {\tt\small Email: \{sibotian, mhzheng\}@tamu.edu}.}%
        \thanks{$^{4}$ Xiao Liang is with Zachry Department of Civil and Environmental Engineering, Texas A\&M University, College Station, TX 77843, USA. {\tt\small Email: xliang@tamu.edu}.}%
		\thanks{$^*$ Correspondence to Minghui Zheng and Xiao Liang.}
}

\maketitle
\begin{abstract}
This paper presents a novel knowledge-informed graph neural planner (KG-Planner) to address the challenge of efficiently planning collision-free motions for robots in high-dimensional spaces, considering both static and dynamic environments involving humans. Unlike traditional motion planners that struggle with finding a balance between efficiency and optimality, the KG-Planner takes a different approach. Instead of relying solely on a neural network or imitating the motions of an oracle planner, our KG-Planner integrates explicit physical knowledge from the workspace. The integration of knowledge has two key aspects: (1) we present an approach to design a graph that can comprehensively model the workspace's compositional structure. The designed graph explicitly incorporates critical elements such as robot joints, obstacles, and their interconnections. This representation allows us to capture the intricate relationships between these elements. (2) We train a Graph Neural Network (GNN) that excels at generating nearly optimal robot motions. In particular, the GNN employs a layer-wise propagation rule to facilitate the exchange and update of information among workspace elements based on their connections. This propagation emphasizes the influence of these elements throughout the planning process. To validate the efficacy and efficiency of our KG-Planner, we conduct extensive experiments in both static and dynamic environments. These experiments include scenarios with and without human workers. The results of our approach are compared against existing methods, showcasing the superior performance of the KG-Planner. A short video introduction of this work is available via this \href{https://zh.engr.tamu.edu/wp-content/uploads/sites/310/2024/03/KGPlanner.mp4}{\textcolor{gray}{\underline{link}}}.
\end{abstract}

\def\abstractname{Note to Practitioners}
\begin{abstract}
This paper was motivated by the problem of human-robot collaboratively working on remanufacturing processes such as disassembly that require human operators and collaborative robots to work closely with each other. The robots need to plan their trajectories efficiently enough to avoid collision with humans and the trajectories need to be short enough to reduce the cycle time. Traditional motion planners usually struggle with finding a balance between efficiency and optimality, which limits wide applications of collaborative robots in remanufacturing systems that are usually less structured than manufacturing systems. This paper suggests a new planning approach that integrates the workspace’s physical information into a graph and leverages deep learning to obtain safe and near-optimal solutions quickly. Experimental studies and observations demonstrated some advantages of this approach including learning capability, efficiency, and optimality, which makes it a great potential approach to be applied to real remanufacturing processes. 
\end{abstract}

\begin{IEEEkeywords}
Motion planning, Graph neural network, Collaborative robot, Human-robot collaboration
\end{IEEEkeywords}

\section{Introduction}

\begin{figure}[!]
	\centering 
	\includegraphics[scale=0.37]{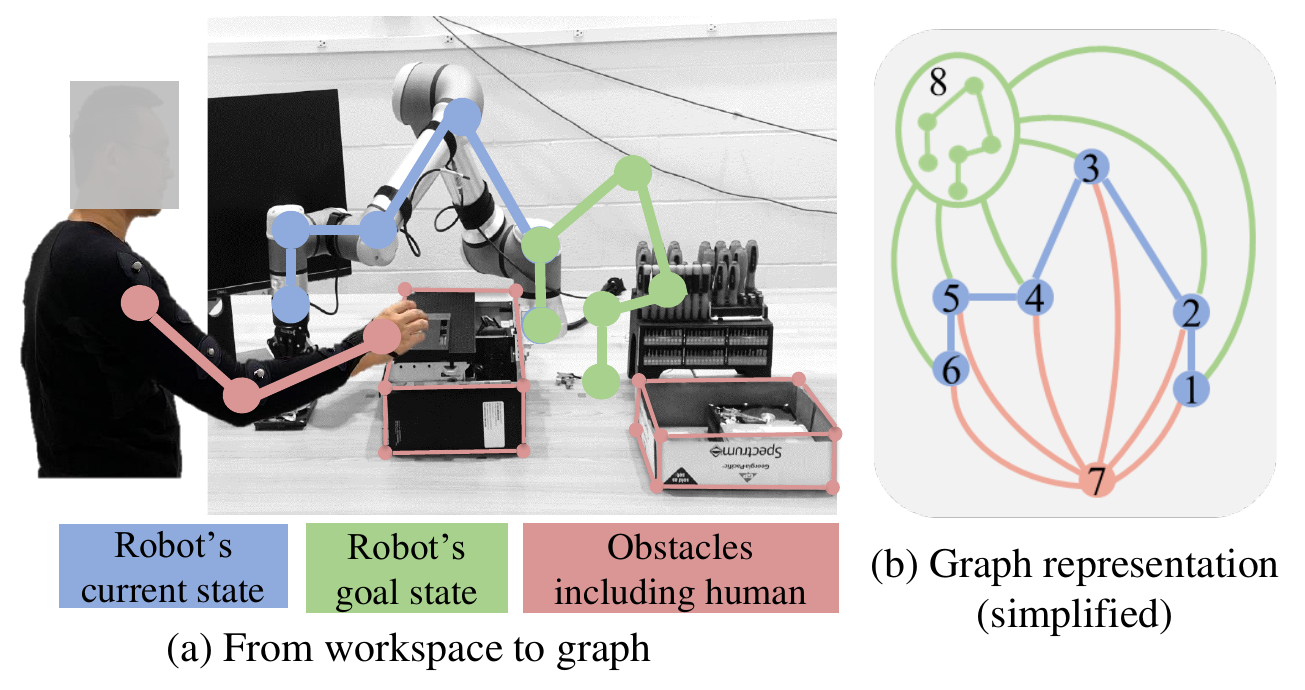}
	\caption{The illustration of the graph representation: (a) The robot's current and desired goal states are represented using blue and green colors, where the nodes indicate the robot joints and the edges indicate the robot links. The static obstacles as well as the human in the workspace are represented using pink color. The nodes of static obstacles indicate their corners, and the nodes of the human indicate the joints of the human arm. (b) The blue nodes denote the joints of the robot's current state. To simplify the graph representation here, we represent the robot's goal state with six small green nodes, and all obstacles with one pink nodes.
 The goal and obstacle nodes are connected with each current joint node since they have effects on the robot's motion generation.
	} \label{fig:introduction}
\end{figure}
Considerable attention has been directed towards the field of robotic motion planning within the context of human-robot collaboration scenarios. The primary objective is to translate high-level collaborative task requirements into precise low-level movement descriptions for robots \cite{berenson2012robot}. An illustrative application of this concept is observed in collaborative disassembly \cite{lee2024review,lee2022robot,lee2022task}, where robots work alongside human operators to disassemble end-of-use products. 
To facilitate a safe and efficient collaboration, robots must rapidly formulate collision-free and near-optimal motion plans within dynamic environments involving human workers \cite{huang2016anticipatory}, a challenge often attributed to the interplay between robot trajectory optimality and computational efficiency \cite{mohanan2018survey,rickert2008balancing,qureshi2015intelligent,qureshi2016potential,liu2023task}.

\begin{figure*}[h]
	\centering 
	\includegraphics[scale=0.32]{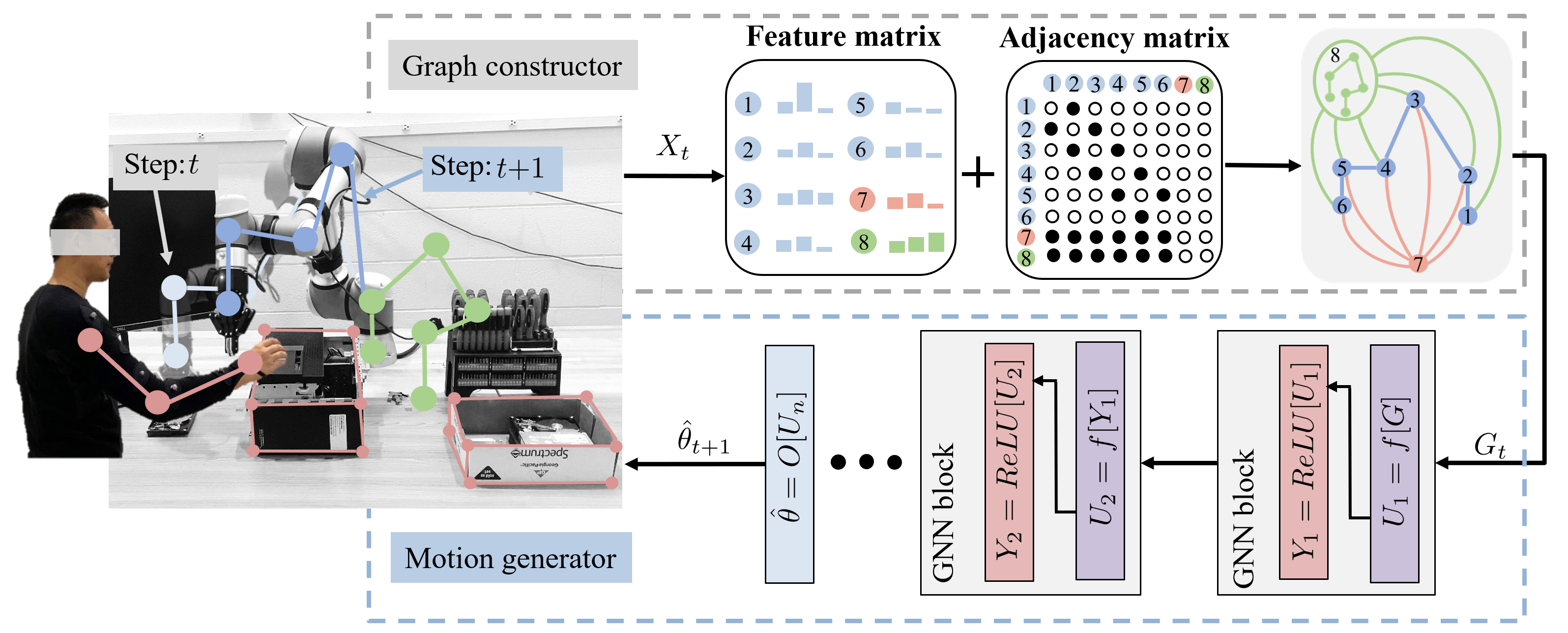}
	\caption{The overview of planning manipulator motion using the proposed KG-Planner: (1) the graph constructor converts the workspace information $X$ of the step $t$ to the feature and adjacency matrices that imply the objects' features and connections, respectively. The black dot in the adjacency matrix indicates the node of the row is directionally connected with the node of the column. (2) The motion generator takes the constructed graph $G$ of the step $t$ as input to generate the manipulator configuration $\hat{\theta}$ of the step $t{+}1$. 
	} \label{fig:overview}
 \vspace{-0.2in}
\end{figure*}

Recently, the utilization of neural networks has emerged as a strategy for effectively managing the trade-off between computational cost and trajectory optimality in motion planning \cite{pfeiffer2017perception,hamandi2019deepmotion,ichter2019robot}. 
For example, existing studies leverage graph neural networks (GNNs) to create policies for quickly generating robot actions and accomplishing specific tasks. Nevertheless, there has been relatively little exploration within robotic motion planning that harnesses the graphical approach to enhance the comprehensibility of motion generation. In other words, there is a gap in how the influence of workspace objects on the generation of robot configurations can be visually and conceptually illustrated. A notable challenge in this pursuit is that the configuration space of a robot constitutes a complex topological entity, particularly in dynamic environments. This complexity renders its explicit representation and visualization through a single graph a challenging endeavor. The difficulty is further magnified when confronting the motion planning of manipulators endowed with a high degree of freedom (DOF). Manipulator joints are interconnected through links, and each joint imparts its unique contribution to parameterizing the overall manipulator configuration. This prompts our conceptualization of neural robot motion planning: instead of blindly mimicking motion patterns from a reference planner, we aspire to explore two critical questions. Firstly, we seek to unravel an optimal approach for a neural planner to consciously leverage the extensive planning knowledge inherent in the workspace. Secondly, we aim to effectively highlight the influences of diverse workspace elements—such as robot joints and obstacles—when orchestrating the generation of secure robot motions.

In our quest to address these inquiries, we deviate from employing a graph to directly represent the robot's configuration space, as done in previous approaches \cite{khan2020graph, zhang2022learning}. Instead, we harness the power of a graph as an intermediary representation to compartmentalize the workspace. This innovative strategy forms the foundation for our proposed solution, termed the Knowledge-Informed Graph Neural Planner (KG-Planner), devised to facilitate the generation of near-optimal manipulator motions during disassembly processes. Our approach entails two crucial steps. Firstly, to harness planning knowledge in an explicit manner, we depict the compositional structure of the workspace through a graph framework. To illustrate, envision a manipulator encompassing 6 joints. As depicted in Fig.\ref{fig:introduction}(a), distinct nodes represent the robot's current state, goal state, and the states of obstacles. These nodes are interconnected by edges. Subsequently, to effectively accentuate the impact of workspace objects, we employ the convolutional operation of a GNN to actively exchange and update information among nodes and their neighboring counterparts. In essence, as we delve into learning robot motions from an oracle planner, the attributes of nodes such as the blue joint node depicted in Fig.\ref{fig:introduction}(b) are continuously refined based on insights from surrounding joint nodes, goal nodes, and obstacle nodes.

In summary, this work introduces significant contributions that can be summarized as follows: 

(1) This study innovatively preserves the structural characteristics of a high-degree-of-freedom manipulator within the data representation, thus aptly capturing the intricate joint dependencies inherent in the system.

(2) The novel KG-Planner is introduced, harnessing the potential of graphs to represent planning knowledge effectively. This planner excels in generating near-optimal manipulator motions while concurrently safeguarding the essential object connectivity.

(3) This work substantiates the effectiveness of the KG-Planner through comprehensive experimental assessments conducted within both static and dynamic environments.

The subsequent sections of this paper are structured as follows. In Section 2, we introduce the related robotic planning works. In Section 3, we lay out the formulation of our motion planning problem using a graph-based approach. Moving to Section 4, we delve into the intricate details of the KG-Planner, encompassing aspects like graph construction, the process of planning knowledge enhancement, network training procedures, and the implementation of online bi-directional planning. The efficacy of the proposed KG-Planner is scrutinized through experimental validation in Section 5. Finally, Section 6
serves as the concluding segment, summarizing the key findings and contributions of this paper.

\section{Related Work}
\subsection{Traditional planning methods}

In recent decades, as the demand for autonomous systems has surged, researchers have investigated a variety of approaches and techniques to address the intricate motion planning problem. Traditional motion planning algorithms can be broadly classified into three main categories: grid-based methods, sampling-based methods, and optimization-based methods.

Grid-based planning algorithms discretize the continuous configuration space into a grid, effectively transforming the space into a finite set of discrete points. Each point on the grid represents a potential configuration of robot. By systematic exploring the grid, the robot could find a feasible path connecting the start node and the goal node, while avoiding obstacles in the environment. The Dijkstra's algorithm \cite{dijkstra2022note} and the $A^*$ algorithm \cite{hart1968formal} are two fundamental grid-based methods. The Dijkstra's algorithm works by iteratively selecting the node with smallest tentative distance from a set of unvisited nodes and updating the distance of its neighbor nodes accordingly. It guarantees to find the shortest path from start to goal if such path exists, but can be computationally expensive. To reduce the number of explored nodes, the $A^*$ algorithm utilizes a heuristic function, such as Euclidean distance, to estimate the cost of reaching the goal from each node and prioritizes nodes that are likely to lead to the shortest path when choosing which nodes to explore. With a greedy search strategy, $A^*$ can find the shortest path more efficiently than the Dijkstra's algorithm. Rooted in these two basic algorithms, many variants \cite{koenig2004lifelong, dolgov2008practical, likhachev2005anytime} have been proposed to address challenges such as considering kinematic feasibility in planning and planning in dynamic environments. In summary, grid-based planning algorithms offer great simplicity, and are widely used in applications such as mobile robot navigation in warehouses. However, as the dimension of the planning space increases, the computational demands tend to grow exponentially. Consequently, grid-based planners become impractical when dealing with motion planning problems of manipulators possessing a high DOF.

Sampling-based planners offer enhanced computational efficiency when tackling planning problems in the high dimensional planning space. Two main types of sampling-based planning algorithms are the probabilistic roadmap (PRM) \cite{kavraki1996probabilistic} and the rapidly exploring random trees (RRT) \cite{lavalle1998rapidly}. The PRM is a multi-query method that constructs a roadmap of the configuration space by sampling feasible configurations and connecting them with collision-free paths, enabling faster query time once the roadmap is built. On the other hand, the RRT is a single-query technique that incrementally expands a tree structure starting from the initial configuration to the goal configuration, making it well-suited for online planning tasks, where the environment is continuously changed. However, these sampling-based methods often struggle to yield optimal solutions, such as the shortest path. Although refinements of classical sampling-based methods, such as RRT* \cite{karaman2011sampling}, informed-RRT* \cite{gammell2014informed}, batch informed trees (BIT*) \cite{gammell2015batch}, and fast marching trees (FMT*) \cite{janson2015fast}, have been proposed to asymptotically approach optimal solutions with improved computational efficiency, they still grapple with the curse of dimensionality. The time taken to search for optimal solutions remains heavily contingent upon the dimensionality of the planning space. What's more, trajectories generated by sampling-based planner usually contain unnecessary nodes, resulting in non-smooth motions without post-processing steps.

Different from previous two categories, optimization-based algorithms formulated the robotic motion planning as an optimization problem, focusing on finding the optimal trajectory with respect to a certain objective function that typically quantifies the quality of a given trajectory, while satisfying several constraints. CHOMP \cite{ratliff2009chomp} optimizes higher-order dynamics and relaxes collision-free feasibility prerequisite, making it suitable for a broad range of inputs. STOMP \cite{kalakrishnan2011stomp} proposes a gradient-free stochastic optimization method that refines the trajectory by sampling noisy trajectories around it and evaluating their possibilities. STOMP dramatically reduces the complexity and makes it applicable to problems with customized and complex objective function. The work in \cite{reynoso2016convex} and \cite{lin2018fast} convert the non-convex problem to a sequence of convex sub-problems in the configuration space, and then solve the subproblems iteratively to get a series motions for manipulators. Marcucci et al. \cite{marcucci2023motion} propose a planner that leverages a convex relaxation to efficiently solve the formulated optimization problem and obtain collision-free robot trajectories. Zimmermann et al. \cite{9001184} develop a multi-level optimization scheme that optimizes the robot grasping locations, the robot configurations, and the robot motions, respectively. Although optimization-based planning methods have the ability to generate smooth trajectories and handle the constraints effectively, such planners have significant drawbacks, such as the local minima problem and high computational complexity.

\subsection{Learning-based planning methods}
Neural networks are being harnessed to either enhance specific components within classical planners or entirely supplant traditional planner pipelines \cite{wang2021survey}.
For example, the integration of neural networks into classical planners has led to advancements such as predicting the probability distribution of the optimal path \cite{wang2020neural}, pinpointing critical sampling points \cite{khan2020graph}, generating distributions of robot configurations pertinent to the task \cite{lehner2018repetition}, as well as predicting and/or streamlining collision checking procedures \cite{liu2022deep,yu2021reducing}. These hybrid approaches focus on bolstering specific planning aspects within classical planners, all with the goal of expediting the motion planning process. Moreover, empirical studies have demonstrated the capacity of purely neural planners to efficiently generate feasible manipulator motions by leveraging learned patterns from motions planned by a reference planner. For instance, Huh et al. \cite{huh2021cost} employed a convolutional neural network to devise collision-free manipulator motions based on point cloud data from the workspace as captured by sensors. Bency et al. \cite{bency2019neural} encoded optimal manipulator motions generated using the $A^*$ algorithm as sequential data, subsequently employing a recurrent neural network to iteratively produce near-optimal manipulator motions. Qureshi et al. \cite{qureshi2020motion} introduced a network-based planner that encapsulates planning information within a latent space via contractive autoencoders, facilitating the generation of subsequent robot configurations towards the goal region. In a similar vein, Li et al. \cite{li2021mpc} pioneered a methodology that utilizes a network to predict a batch of forthcoming robot states. This prediction is subsequently fine-tuned through model predictive control, ensuring alignment with desired constraints.

Despite the advantages exhibited by the aforementioned neural planners in motion planning for robots, they still grapple with certain limitations. Primarily, it is important to acknowledge that point cloud data encompasses not only pertinent planning information, such as obstacle locations and robot states, but also extraneous data, including background details from the workspace. Consequently, applying convolutions to point cloud data can lead to redundant computations \cite{zhang2019three, chen2019drop, liu2021dynamic}. Furthermore, as pointed out in \cite{li2021directed}, a simple transformation of learning-relevant information into sequential data or latent representations may fail to capture the inherent connectivity among objects \cite{liu2022dynamic, park2018sequence}. For instance, successful robot motion planning necessitates a thorough understanding of the states of other objects within the workspace. However, current neural planners often overlook the intricate ways in which other objects influence the generation of robot motion. Consequently, these planners lack the ability to account for crucial object dependencies.

For efficient data representation and improved object connectivity, the conversion of objects into a graph format emerges as a promising solution, as graphs inherently depict sets of objects along with their interconnections \cite{sanchez-lengeling2021a}. The realm of research has extensively embraced the utilization of GNNs for learning policies through expert demonstrations, ultimately allowing for the imitation of robotic actions like box picking and block stacking. For instance, Ding et al. \cite{ding2022visual} introduced a concept where all objects within a scene are designated as nodes within a graph. The interrelationships among these nodes, based on their manipulation interactions, are established as edges. This constructed graph of relationships aids the robot in deducing the optimal grasping sequence for objects. Similarly, Lin et al. \cite{lin2022efficient} devised a graph-based policy, training it to discern the appropriate nodes indicating the subsequent object to be picked and the desired placement location. Moreover, Huang et al. \cite{huang2019neural} presented a novel neural network approach that employs graphs to capture the compositional aspects of visual demonstrations. This approach has successfully enabled the imitation of robot actions, such as picking and placing, through the learned graph-based representations.

\section{Problem Formulation}
In this section, we introduce the notations and definitions of variables used in this paper, and briefly present the formulation of our graph neural planning problem.

Let $\Theta$ be the manipulator configuration space, $\Theta_{obs}$ be the obstacle space, and $\Theta_{free}$ be the obstacle-free space, where $\Theta_{free}=\Theta \setminus \Theta_{obs}$. We denote the manipulator configuration as 
 $\theta \subset \mathbb{R}^{q}$, where $q$ is the dimension of the configuration space. Note that the configuration planned by our KG-Planner in this paper is denoted with $\wedge$. 
Given manipulator start and goal configurations in a workspace, our work aims to use a graph-based planner $\Delta(\bullet)$ learning from an oracle planner to plan a series of near-optimal manipulator motions $[\hat{\theta}_1,\dots,\hat{\theta}_t,\dots,\hat{\theta}_T]$, where $\hat{\theta}_t \in \Theta_{free}$ and $T$ is the step horizon. 

The planner $\Delta(\bullet)$ comprises of two components: a graph constructor $\Delta_{con}(\bullet)$ and a motion generator $\Delta_{gen}(\bullet)$. As shown in Fig.~\ref{fig:overview}, $\Delta_{con}(\bullet)$ extracts the workspace information $X$ 
as two matrices describing the feature and structure knowledge of the workspace to construct a graph $G$, i.e., $G=\Delta_{con}(X)$. The workspace actually has two categories of information: pre-known information and real-time information. The pre-known information includes the robot's start state, the robot's goal state, and the positions of static obstacles. These details are ascertainable before the planning process begins.
Considering that the static obstacles in this study are predominantly rectangular, it is feasible to determine the corner position of each static obstacle, and define the corresponding nodes.
Consequently, the joint values of the robot and the corner positions of static obstacles can be easily converted into the features of nodes. Meanwhile, the real-time information includes the positions of dynamic obstacles obtained by a Vicon motion tracking system. 
The real-time positions of dynamic obstacles, such as human arm, are also converted into the features of the corresponding nodes. Furthermore, 
the adjacency matrix is a binary matrix that can be manually defined to indicate relationships between nodes. For example, the connection between robot joint nodes and the connection between the robot joint node and the obstacle node.
In particular, the information used for training is denoted as $X_{seen}$, and the information used for testing is denoted as $X_{unseen}$. The motion generator $\Delta_{gen}(\bullet)$ is trained using motions collected from an oracle planner based on $X_{seen}$, and takes $G$ as the input to generate the next manipulator configuration $\hat{\theta}$ toward the goal, i.e.,  $\hat{\theta}=\Delta_{gen}(G)$.  
Eventually, given the unseen workspace information $X_{unseen}$, by employing bi-directional planning strategy, the well-learned planner can recursively plan collision-free motions for robots, and efficiently lead the manipulator to reach the goal.

\section{Graph Neural Planner}
This section presents the details of our KG-Planner. We (1) explicitly preserve the workspace information using graph representation, (2) efficiently utilize the planning knowledge to update the nodes of the graph, (3) train a GNN offline, and (4) implement the well-trained GNN on online bi-directional motion planning.

\subsection{From workspace to graph}
Manipulator motions are planned based on workspace states. The input of our KG-Planner does not consist of the image observation of the workspace or the point cloud representation. Instead, the input of our KG-Planner directly comprises the pre-known information including the joint values of the robot states and the positions of the static obstacle, as well as the real-time information including the positions of the human arm. Considering that each component in the workspace has its own features, the graph constructor $\Delta_{con}(\bullet)$ extracts the key features of the components, and represents each of them using node $v$ associated with a feature vector $c(v)$. For example, to preserve the manipulator structure attributes enhancing the joint dependency, the constructor denotes each joint of the manipulator as a node, and successively connect them with edges, which is illustrated in Fig.~\ref{fig:introduction}(a). Furthermore, to highlight the component connectivity, as shown in Fig.~\ref{fig:introduction}(b), each joint node is connected with both the obstacle nodes and the goal nodes such that the obstacle and goal states would have effects on the manipulator's joints during planning. 

The node feature extraction comprises three cases. The first case involves extracting the start and goal states of the robot. As these details are ascertainable prior to the planning stage, we simply convert the joint values of the robot's start and goal configurations into the features of the robot-state-related nodes. This choice of utilizing joint values, as opposed to joint positions, offers advantages in addressing the inconsistent link length issue caused by bringing the relative translation into training. Similarly, in the second case, which pertains to the extraction of static obstacles, we possess foreknowledge of the obstacles' positions. Considering the static obstacles in this study are predominantly rectangular, we select the corners of each static obstacle and convert the positions of the corners into the features of the static-obstacle-related nodes. The third case involves the extraction of dynamic obstacles, with the joints of the human arm serving as dynamic-obstacle-related nodes. By employing a Vicon motion tracking system, we obtain real-time positional data for the human arm and subsequently convert the real-time joint positions of human arm into the features of the dynamic-obstacle-related nodes. What's more, we do not include embedding layers to these features; instead, we utilize zero padding to match the dimension between each node. Note that the green nodes should have the same number as the blue nodes, and the total number of pink nodes depends on the number and shapes of different obstacles. To simplify the representation, we only show six small green nodes and one pink node in Fig.~\ref{fig:introduction}(b).

In summary, the workspace contains feature knowledge (e.g., a feature matrix) and connection knowledge (e.g., an adjacency matrix). 
We use a graph $G=\{V,E\}$ to capture the planning knowledge of the workspace, where a set of nodes $V{=}[v_1,\dots,v_m,\dots,v_M]$ indicates total $M$ objects in the workspace, and a set of edges $E$ implies the connection between nodes.

\subsection{Planning knowledge update}
The previous subsection briefly presents how to construct a graph from a workspace. The graph description of the feature and adjacent matrices preserves the property of permutation invariances \cite{gama2020stability}. Such a property ensures that the graph can be operated without changing the object connectivity.
To efficiently utilize the knowledge stored in nodes as well as the object connectivity, we use the convolution operation of GNN to update the node knowledge.

\begin{figure}[h]
	\centering 
	\includegraphics[scale=0.47]{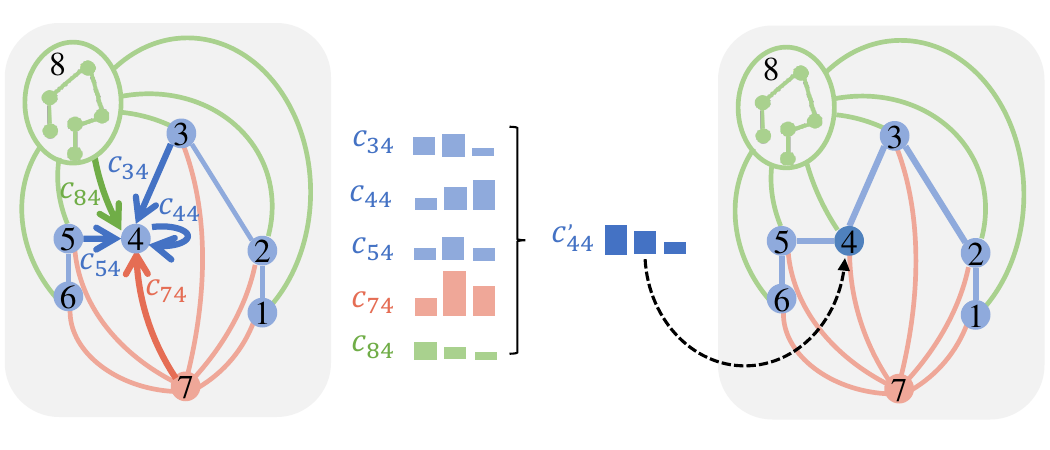}
	\caption{The update of node embedding in the GNN model: the joint node 4 receives the planning knowledge from its neighbors, $c_{34}$ means the feature vector sent from the joint node 3 to the joint node 4, and the node embedding update happens on all nodes simultaneously.
	} \label{fig:update_node}
\end{figure}

GNN shows the great capability of operating on graphs. The GNN layer updates each node embedding based on its neighbors (i.e., directly connected nodes using edges). We denote $N_{v_m}$ as the set of neighbor nodes used to update the embedding of $v_m$. The node embedding updates using the following equation:

\begin{equation}
    h^k_{v_m}=\sigma\left(f^k_w(h^{k-1}_{v_m}, \{h^{k-1}_{v_i}\}_{i \in N_{v_m}})\right)
    \label{eqn:node_update}
\end{equation}
where $\sigma$ is a $ReLU$ activation function, $h^k_{v_m}$ is the updated embedding of node $v_m$ in hidden layer $k$, and $f^k_w$ indicates the knowledge update function with weights $w$. Fig.~\ref{fig:update_node} presents an example of the embedding update, where the joint node 4 of the robot's current state collects the embeddings from its neighbors to update itself based on Eq.~(\ref{eqn:node_update}). In this case, the states of obstacles, all six small green nodes indicating the manipulator goal, and the 3rd and 5th manipulator joint would all influence the updated embedding of the joint node 4.

We simultaneously update all nodes in layer $k$ using a layer-wise propagation rule \cite{kipf2016semi}: 

\begin{equation}
H^{k}=\sigma\left(\tilde{D}^{-\frac{1}{2}} \tilde{A} \tilde{D}^{-\frac{1}{2}} H^{{k-1}} W^{k-1}\right)
\end{equation}
where $H^{k}$ is the updated layer embedding including all nodes in matrix form, $\tilde{A}=A+I_M$ is the binary adjacency matrix of the graph $G$ with added an identity matrix $I_M$, $\tilde{D}_{mm}=\sum_i\tilde{A}_{mi}$ indicates a degree matrix, and $W^{k-1}$ is a learning weight matrix. 
In addition, considering that the embedding update is not needed for all nodes, i.e., the obstacle and goal nodes, we specify the adjacency matrix $A$ to ensure a certain directional update between nodes. For example, the black dot
in the adjacency matrix shown in Fig.~\ref{fig:overview} implies the embedding only updates from the nodes of the row to the nodes of the column.

By stacking multiple hidden layers, even if the nodes are not directly connected, they are also capable of exchanging planning knowledge. For example, the joint node 1 and 6 are not directly linked by edges shown in Fig.~\ref{fig:update_node}, they would also influence each other's updated embeddings after several layer-wise propagations. Eventually, the next manipulator configuration $\hat{\theta}$ is predicted based on all updated nodes using the following equation: 

\begin{equation}
    \hat{\theta}=O_w(\textstyle \sum_{v_m \in G}h^k_{v_m})
\end{equation}
where $O_w$ stands for the output layer with weights $w$. Note that in the robotic motion planning problem of this study, it is assumed that the start and goal states of the robot are pre-known, and the primary objective is to find a sequence of collision-free motions that connect the specified start and goal robot configurations. Therefore, our KG-Planner is trained and tested based on a goal-oriented manner.

\subsection{Network training}
This subsection introduces that our KG-Planner learns from an oracle planner to generate near-optimal manipulator motions. We borrow the prediction strategy from \cite{qureshi2020motion}, which uses the workspace state as the planning information to predict one-step ahead for the robot.  

Let the robot motion sequences from an oracle planner $\mathcal{E}=\{\epsilon_1,\dots,\epsilon_p,\dots,\epsilon_P\}$ be the training dataset, where $\epsilon=[\theta_1,\dots,\theta_T]$ is an optimal manipulator motion, bridging the given start and goal configurations. The training of GNN iteratively updates the parameters of our KG-Planner according to the loss of the training sample batch. Especially, instead of blindly intercepting the manipulator configuration $\theta$ from the whole training dataset $\mathcal{E}$, we construct the traning batch using the manipulator motion $\epsilon$, thus enhancing the motion integrality during training. The loss function is defined using the following equation:

\begin{equation}
l_{\text {KG}}=\frac{1}{N_p} \sum_p^{N_p} \sum_{t=0}^{T_p-1}\left\|\theta_{p, t+1}-\hat{\theta}_{p, t+1}\right\|^2
\end{equation}
where $N_p$ is the batch size, and $T_p$ is the length of the $p$th motion.

\subsection{Online bi-directional motion planning}
This subsection presents the details of online motion planning based on the well-learned KG-Planner. At test time, a straightforward way of online planning is to predict the next manipulator configuration using KG-Planner, and recursively replace the manipulator state with the predicted one until the prediction reaches the goal. 
 However, the predicted configuration highly relies on the prediction from the previous iteration due to such a one-step-ahead prediction strategy. The prediction error would accumulate throughout the entire manipulator motion sequence, which may lead the manipulator to a position that is far away from the goal. Therefore, to have more robust online planning, we employ a bi-directional planning strategy, which plans the manipulator motion from both start and goal configurations simultaneously.  

\begin{figure}[h]
	\centering 
	\includegraphics[scale=0.44]{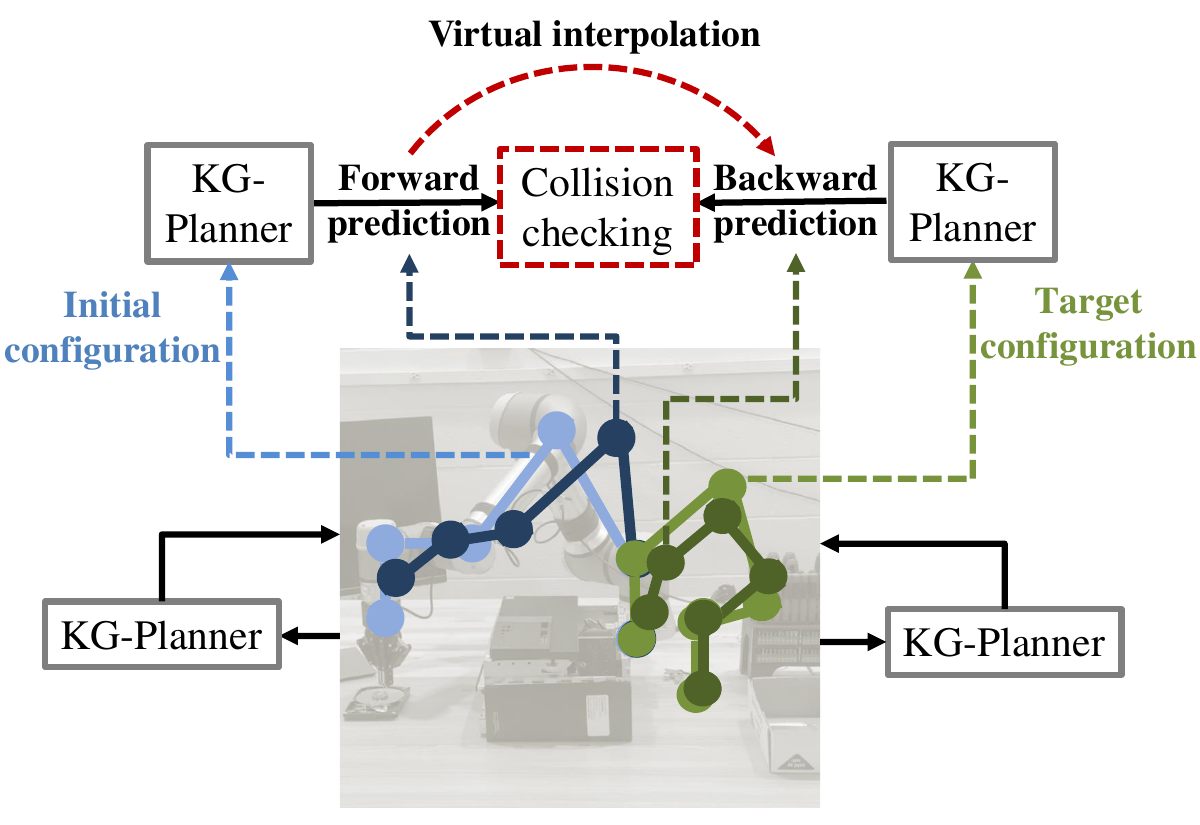}
	\caption{The illustration of KG-Planner-based bi-directional planning: The KG-Planner takes the initial and target configurations as inputs to provide forward and backward predictions simultaneously. Linearly virtual interpolations are generated between two predictions, which try to directly connect two planning branches. If there is a collision between the connected planning branches and the environment, the predictions become the new inputs of the KG-Planner. We do such bi-directional planning recursively until a collision-free path is found.  
	} \label{fig:Bi}
\vspace{-0.2in}
\end{figure}

\begin{figure*}[h]
	\centering 
	\includegraphics[scale=0.53]{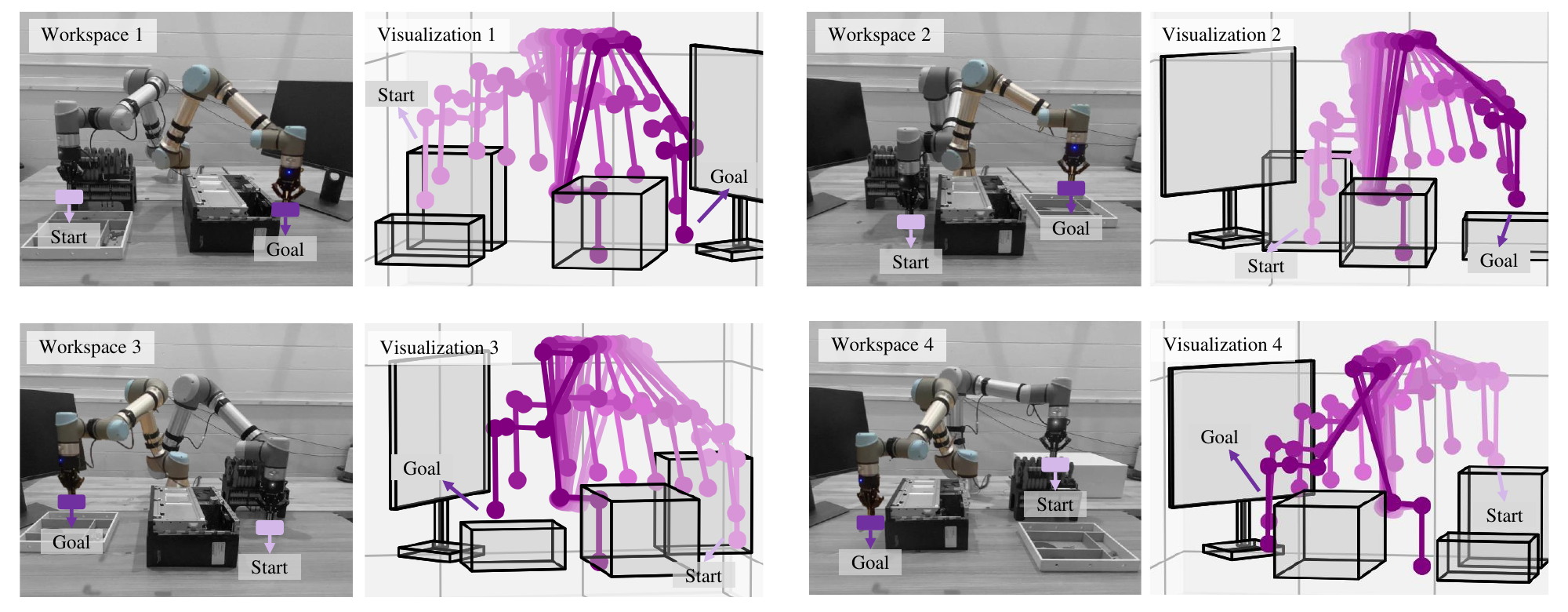}
	\caption{Experimental planning results in different environments: given random start and target manipulator configurations in the selected four environments, our bi-directional KG-Planner plans near-optimal motions.}
 \label{fig:KG_Planner_demos}
\vspace{-0.1in}
\end{figure*}

Fig.~\ref{fig:Bi} illustrates our bi-directional planning using the KG-Planner. The planning separately generates two branches from forward and backward directions. 
To guarantee safety, we have two requirements of collision checking in each planning iteration: (1) we directly check if the two predictions obtained from the KG-Planner are feasible or not. (2) we first virtually connect the forward and backward predictions, then generate several linearly interpolated configurations between them, and finally check if there is any collision between obstacles and these virtual interpolations. The planning is terminated when two requirements are satisfied, and a complete manipulator motion is generated by stitching two branches together. 

\begin{algorithm}[h]
    $\bullet$  $\hat{\Theta}^{a \rightarrow b}$ $\leftarrow$ \textit{interpolation}($\theta_{start}$, $\theta_{goal}$)\\
	$\bullet$  $\hat{\epsilon}^a$ $\leftarrow$ \textit{add}($\theta_{start}$), $\hat{\epsilon}^b$ $\leftarrow$ \textit{add}($\theta_{goal}$)\\
    $\bullet$ Check \textit{collision}($\hat{\Theta}^{a \rightarrow b}$)\\
    \While{collision is True}
    {
    $\bullet$ $G^a$ $\leftarrow$ $\Delta_{con}(X^a)$, $G^b$ $\leftarrow$ $\Delta_{con}(X^b)$\\
    $\bullet$ $\hat{\theta}^a \leftarrow \Delta_{gen}(G^a)$, $\hat{\theta}^b \leftarrow \Delta_{gen}(G^b)$ \\
    $\bullet$  Check \textit{feasible}($\hat{\theta}^a$, $\hat{\theta}^b$)\\
    \eIf{feasible is True \textbf{and} step $<$ $\delta$}
                {$\bullet$ $\hat{\Theta}^{a \rightarrow b}$ $\leftarrow$ \textit{interpolation}($\hat{\theta}^a$, $\hat{\theta}^b$)\\
                $\bullet$ $\hat{\epsilon}^a$ $\leftarrow$ \textit{add}($\hat{\theta}^a$), $\hat{\epsilon}^b$ $\leftarrow$ \textit{add}($\hat{\theta}^b$)\\
                $\bullet$ Check \textit{collision}($\hat{\Theta}^{a \rightarrow b}$)\\
                $\bullet$ $X^a \leftarrow \hat{\theta}^a$,$X^b \leftarrow \hat{\theta}^b$
                }
                {$\bullet$ \textbf{return} $\emptyset$}
    $\bullet$ Increment \textit{step}
    }
    $\bullet$ \textbf{return} \textit{stitch}($\hat{\epsilon}^a$, $\hat{\epsilon}^b$)
    \caption{Online Bi-direction Planning}
	\label{bi-online}
\end{algorithm}   

The bi-directional planning procedure is outlined in \textbf{Algorithm~\ref{bi-online}}. 
The forward branch $\hat{\epsilon}^a$ plans the manipulator motion from start to goal, and the backward branch $\hat{\epsilon}^b$ plans the manipulator motion from goal to start. 
The future planning knowledge of the workspace is recursively updated using $\Delta_{con}(\bullet)$ and $\Delta_{gen}(\bullet)$. After every branch expansion, we try to directly connect two branches based on the linearly virtual interpolations such that the joint change effort of the manipulator can be reduced. Note that in cases where both the position and the orientation of the goal state are unknown, our KG-Planner cannot plan a trajectory towards the goal. Additionally, when the position of the goal state is known but the orientation is unknown, we can employ inverse kinematics to determine a feasible goal configuration for the robot. Subsequently, we can proceed with the same planning procedure to obtain the robot trajectory.

\begin{figure*}[h]
	\centering 
	\includegraphics[scale=0.62]{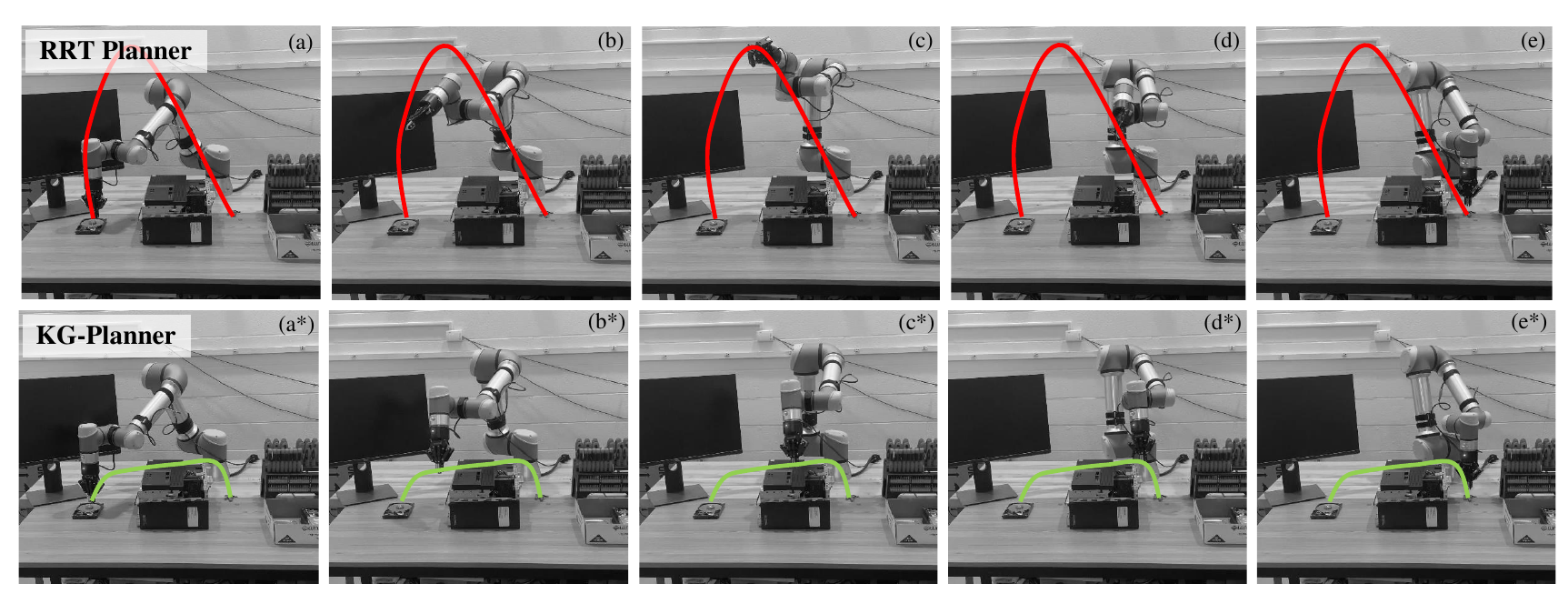}
	\caption{The experimental comparison tests: The top and bottom rows show the planning using the RRT planner and our bi-directional KG-Planner, respectively. They share the same start and goal configurations. We denote the two different end-effector paths using red and green colors. The \textbf{\underline{\smash{experimental video}}} is available via this \href{https://zh.engr.tamu.edu/wp-content/uploads/sites/310/2024/03/KGPlanner.mp4}{\underline{link}}.}
	 \label{fig:comparison_1}
\end{figure*}

\begin{table*}[h]
	\centering
	\begin{tabular}{cccc}
\toprule[1.5pt]		Planner  & Path cost (m) & Planning time (s) & Success rate
\\\toprule[1.5pt]
        RRT  & 1.695 $\pm$ 0.725 & 0.335 $\pm$ 0.650 & 93.5\%\\ 

		KG-Planner A & 1.362 $\pm$ 0.227 & 0.476 $\pm$ 0.135  & 92.9\% \\
		
		KG-Planner B  & \textbf{1.023 $\pm$ 0.130} &  \textbf{0.197 $\pm$ 0.053} & 88.1\% \\
		
		RRT*($\pm$10\% of KG-Planner B cost)   & 1.078 $\pm$ 0.138 & 3.700 $\pm$ 2.421 & 70.2\%\\
  
        RRT*($\pm$30\% of KG-Planner B cost)   & 1.151 $\pm$ 0.162 & 2.072 $\pm$ 1.372 & 75.5\%\\

        BFMT*($\pm$10\% of KG-Planner B cost)   & 1.082 $\pm$ 0.137 & 0.541 $\pm$ 0.148 & 91.1\%\\

        BFMT*($\pm$30\% of KG-Planner B cost)   & 1.167 $\pm$ 0.173 & 0.417 $\pm$ 0.131 & 96.5\%\\
		\toprule[1.5pt]
	\end{tabular}
	\vspace{5pt}
	\caption{Summary of planning results: We assume that the result data fit normal distributions, and present them using the mean and standard deviation. KG-Planner A denotes the single-directional KG-Planner, and KG-Planner B stands for the bi-directional KG-Planner. Considering that the optimality of the motions planned by RRT* and BFMT* is based on the given planning time, we terminate the plannings of RRT* and BFMT* once the path cost is within a certain range of the one planned by KG-Planner B. Note that the planning time using RRT may not fit a normal distribution, but we still put such data here to have a better comparison with the planning time using other planners.
 In general, the bi-directional KG-Planner outperforms the other planners regarding the path cost and planning time, while maintaining an acceptable success rate.} 
	\label{tab:comparison_detail}
\vspace{-0.1in}
\end{table*}

\begin{figure*}[h]
	\centering 
	\includegraphics[scale=0.5]{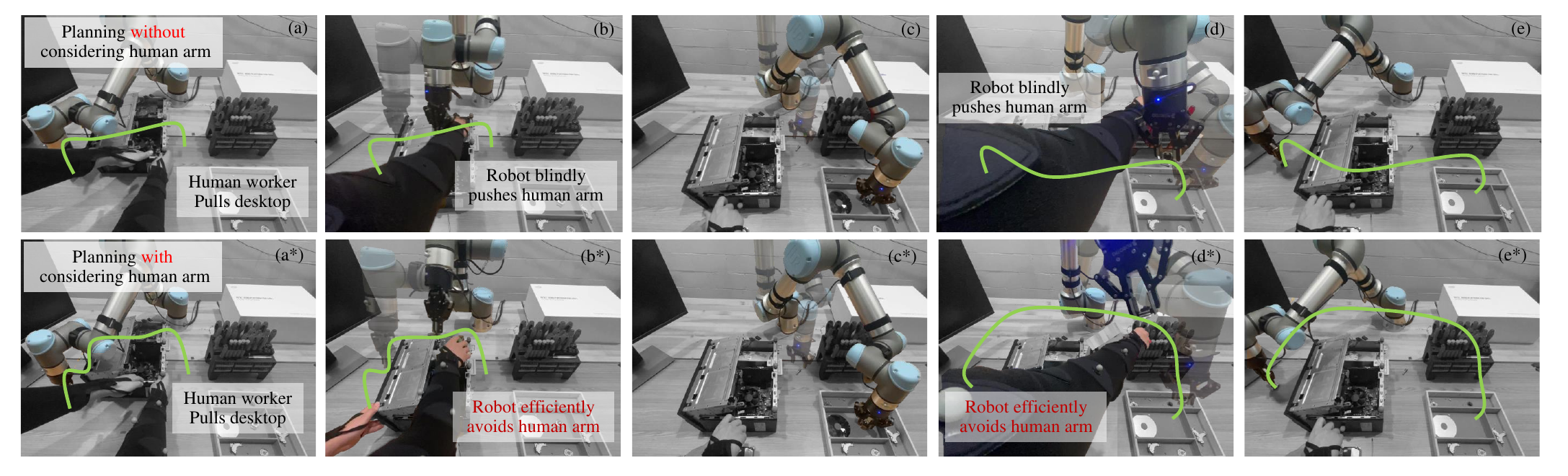}
	\caption{The experimental tests without and with considering the human arm: The top row shows the task execution without considering the human arm. It contains tow planning scenarios: scenario A (i.e., carrying a component from left to right, and shown in (a) and (b)) and scenario B (i.e., carrying a component from right to left, and shown in (d) and (e)). The sub-figure (c) shows the transition between two planning scenarios. The transparent robot arms in sub-figures indicate the past movements of the robot. The robot follows the originally planned motion and blindly pushes the human arm. 
 The bottom row presents the planning with the same initial and target configurations with the consideration of the human worker's real-time motions, where the robot effectively and efficiently re-plans its motions when a possible collision is detected. The Vicon system tracks the positions of the human arm and desktop based on the attached gray markers in real time. 
The \textbf{\underline{\smash{experimental video}}} is available via this \href{https://zh.engr.tamu.edu/wp-content/uploads/sites/310/2024/03/KGPlanner.mp4}{\underline{link}}.\label{fig:comparison_2}}
\vspace{-0.2in}
\end{figure*}

\section{Experimental Tests}
This section validates the efficacy of our KG-Planner in disassembly processes. We collect the motions from an oracle planner using simulation, and experimentally implement the online bi-directional planning in both static and dynamic environments.

\subsection{Tests in static environments}
\subsubsection{Data acquisition and network training}

We simulate a disassembly scenario in an open-source motion planning platform MoveIt. We use RRT* as the oracle planner to generate the optimal motion used for learning. A collaborative robot UR5e with 6-DOF is employed to conduct all experimental tests. The obstacles in the workspace include a monitor, a desktop, a screwdriver box, and a disassembly container. 
We create 12 different workspaces, each containing 800 planning scenarios with randomly generated start and goal configurations. 90\% of these scenarios are allocated for training purpose, while the remaining 10\% are reserved for testing. 

\subsubsection{Planning results and discussions}
Fig.~\ref{fig:KG_Planner_demos} illustrates the planning results using the bi-directional KG-Planner in 4 out of 12 workspaces. The successful planning based on different scenarios shows that our KG-Planner can be employed for the manipulator in disassembly tasks, e.g., carrying disassembled components to the container or taking a screwdriver to a desired location.  
We also compare our approach with the classical planner i.e., RRT, the oracle planner used for training, i.e., RRT*, and the advanced planner, i.e., bi-directional FMT* (BFMT*). To have a fair comparison, all planners used for comparison studies come from the open motion planning library (OMPL).  
Fig.~\ref{fig:comparison_1} illustrates the experimental comparison between our planner and the RRT planner. The manipulator aims to carry a disassembled component from a start configuration to a goal configuration, and the planners need to provide collision-free motions for the manipulator. The green and red curves are the planned end-effector's path using our bi-directional KG-Planner and RRT, respectively.

To better demonstrate the planning results, the above-mentioned methods share the same environments during the planning, and we define three measurements: (1) the path cost denotes the trajectory length of the manipulator's end-effector, which implies the trajectory optimality. (2) the planning time denotes the planning duration based on a random pair of start and goal configurations, which implies the approach efficiency. (3) the success rate denotes the percentage of the successful planning instances over the total planning attempts, which implies the approach efficiency. The planning of the RRT planner terminates once it finds a collision-free robot trajectory. For RRT* and BFMT*, where the trajectory could potentially be more optimal with increased planning time, termination criterion is based on two rules. The first rule terminates the planning once the path cost is within a certain percentage of the one planned our KG-Planner. This enables a fair comparison of computational time and success rate with nearly identical path costs. The second rule terminates the planning if the planning time exceeds the maximum planning time, set at 8 seconds, acknowledging the impracticality of prolonged planning time in human-robot collaboration scenarios. Additionally, the 10\% of generated planning scenarios (i.e., 960 scenarios) is used to evaluate the planning results in terms of path cost, planning time, and success rate. The workspaces are seen, while the pairs of start and goal configurations are unseen for the KG-Planner.

The average results are shown in \textbf{TABLE}~\ref{tab:comparison_detail}, where KG-Planners A and B indicate single-directional and bi-directional KG-Planners, respectively. Note that our KG-Planner significantly outperforms the oracle planner (i.e., RRT*) in terms of success rate. Planning failures with RRT* can be categorized into two primary cases. The first case occurs when the RRT* planner is unable to identify a collision-free path due to its inherent random sampling property, and this scenario constitutes a relatively small proportion of planning failures.
The second case, which constitutes the majority of planning failures, involving the planning duration surpassing the maximum allowable planning time. 
We terminate the planning of RRT* until the corresponding path cost is within a certain percentage of the one planned by our KG-Planner. However, during the bi-directional planning, the virtual interpolations in our KG-Planner attempt to directly connect two planning branches in each iteration, which would eliminate the unnecessary planning and further reduce the associated path cost. Therefore, RRT* is difficult to find a sufficiently short path within the limited time duration.

Furthermore, we keep the same types and total number of obstacles, and change the locations of the desktop, the screwdriver box, the monitor, and the container to construct three unseen workspaces which are different from the 12 workspaces used for training. Random pairs of start and goal robot configurations are given to our KG-Planner. The planner is required to find collision-free robot motions to connect the given start and goal configurations. Such tests are used to validate the generalization of our bi-directional KG-Planner. The success rate of planning is shown in \textbf{TABLE}~\ref{tab:unseen}.

\begin{table}[h]
	\centering
	\begin{tabular}{cc}
\toprule[1.5pt]		Unseen workspace & Success rate
\\\toprule[1.5pt]
         A  & 94.0\% (376/400)\\ 

		 B   & 88.0\% (352/400) \\
		
		 C   & 81.0\% (324/400)\\
		\toprule[1.5pt]
		Overall    & 87.6\%\\
  
		\toprule[1.5pt]
	\end{tabular}
	\vspace{5pt}
	\caption{Planning results for unseen workspaces: A, B, and C stand for three different workspaces in this table, and we randomly generate 400 planning scenarios for each unseen workspace.}
	\label{tab:unseen}
 \vspace{-0.2in}
\end{table}

The above-mentioned results indicate a few points: (1) The bi-directional KG-Planner is capable of providing near-optimal manipulator motions since KG-Planner learns from an oracle planner and the generated virtual interpolations keep trying to connect the forward and backward planning branches. (2) Our approach can efficiently plan collision-free manipulator motions due to the short inference time of the GNN model. (3) Compared to RRT and BFMT*, using KG-Planner to obtain manipulator motions has less success rate. Additionally, requiring both the forward and backward predictions to be collision-free (i.e., KG-Planner B), further decreases the planning success rate. (4) The bi-directional KG-Planner shows the great capability of generalization to similar but unseen workspaces.

\subsection{Tests in dynamic environments}
\subsubsection{Data acquisition and network training}
The previous subsection presents the efficacy of the bi-directional KG-Planner in the static environment. Due to the short planning time, our approach also has the applicable potential of re-planning the manipulator motion in real-time to avoid the human worker. To generate training data, we import the movement of the human arm captured by the Vicon motion capture system into one workspace. The human arm is built using multiple geometric models in MoveIt and treated as an additional obstacle. 
We employ RRT* to generate collision-free 12000 manipulator motions in the workspace involving human motion.
For the data generated from the dynamic environment, 90\% is used for training, and the remaining is used for testing.

\begin{figure}[!htbp]
	\centering 
	\subfigure[Planning scenario A: the joint angles of the UR5e in the planning without considering the human worker, the planned robot motion shows in sub-figures (a) and (b) of Fig.~\ref{fig:comparison_2}, the orange shadow indicates the virtual interpolation part, and the orange dot stands for the start and end moments that the robot moves following the interpolations.]{
	\includegraphics[scale=0.28]{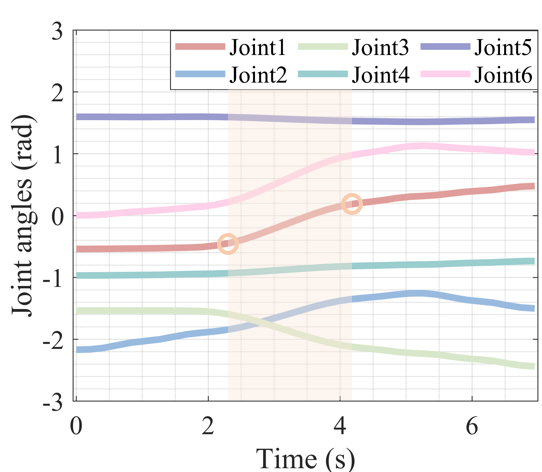}
	\label{fig:joint_without_arm_a}
	}
\hfill
 \subfigure[Planning scenario B: the joint angles of the UR5e in the planning without considering the human worker, the planned robot motion shows in sub-figures (d) and (e) of Fig.~\ref{fig:comparison_2}, the orange shadow indicates the virtual interpolation part, and the orange dot stands for the start and end moments that the robot moves following the interpolations.]{
	\includegraphics[scale=0.28]{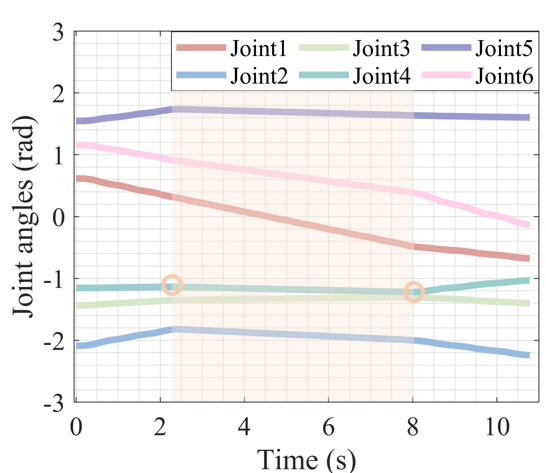}
	\label{fig:joint_without_arm_b}
	}
	\subfigure[Planning scenario A: the joint angles of the UR5e in the planning with considering the human worker, the planned robot motion shows in sub-figures (a*) and (b*) of Fig.~\ref{fig:comparison_2}, and the red shadows indicate the re-planning parts.]{
	\includegraphics[scale=0.28]{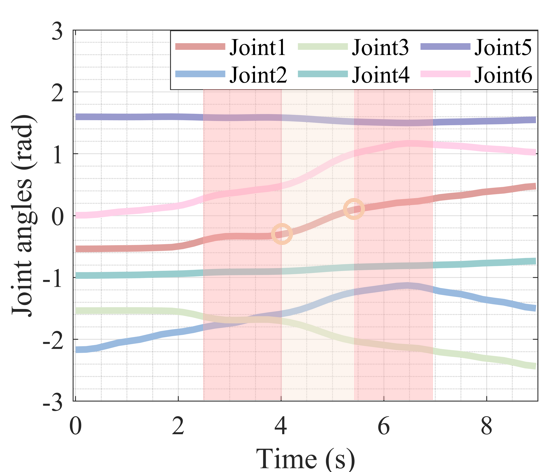}
    \label{fig:joint_with_arm_a}
	}
 \hfill
\subfigure[Planning scenario B: the joint angles of the UR5e in the planning with considering the human worker, the planned robot motion shows in sub-figures (d*) and (e*) of Fig.~\ref{fig:comparison_2}, and the red shadows indicate the re-planning parts.]{
	\includegraphics[scale=0.28]{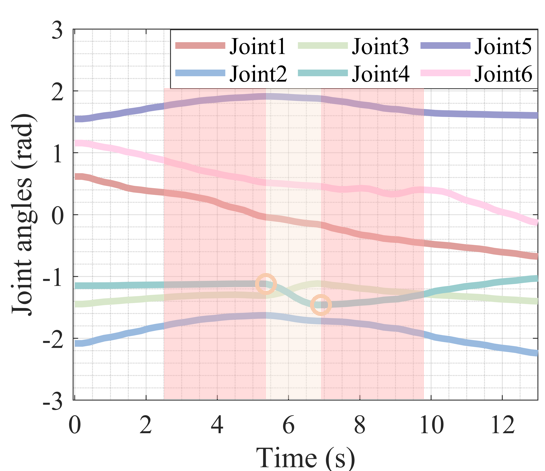}
    \label{fig:joint_with_arm_b}
	}
 
	\caption{The joint angle comparison between two planning scenarios.}
	\label{fig:comparison_angle}
\vspace{-0.25in}
\end{figure}

\subsubsection{Planning results and discussions}
When a manipulator conducts a disassembly task, our planner first plans an initial collision-free motion for the manipulator and continuously checks if a new motion is needed to be re-planned based on the current workspace state such that the manipulator can avoid a sudden entry of the human worker. 
The simulated planning success rate in such a dynamic environment is 79.4\%.

 Fig.~\ref{fig:comparison_2} presents the task executions without and with considering the human worker. The sub-figures (a) and (b) show the planning scenario A: carrying a disassembled component from left to right, and the sub-figures (d) and (e) indicate the planning scenario B: carrying a disassembled component from right to left. Additionally, the sub-figure (c) shows the transition between two planning scenarios. Note that the real-time motions of the desktop and human worker are both tracked and sent to our planner.
 As shown in the top row of Fig.~\ref{fig:comparison_2}, the manipulator directly collides with the human worker and follows the originally planned motion pushing the arm. On the contrary, the bottom row of Fig.~\ref{fig:comparison_2} shows that once the human arm collides with the originally planned manipulator motion, our planner immediately re-plans the manipulator's motion and enables the manipulator to efficiently avoids the potential collision. Fig.~\ref{fig:comparison_angle} presents the joint angles of the manipulator in the two task executions. 
 The orange shadow stands for the virtual interpolation part connecting the forward and backward planning, the orange dot represents the start and end moments that the robot moves following the interpolations, and the red shadow indicates the re-planning part caused by the human worker movements. 

\vspace{-0.05in}
\section{Conclusions} 

This paper introduces a novel graph-based neural planner designed to adeptly generate collision-free motions that are nearly optimal for collaborative robots. Our KG-Planner capitalizes on the intricate interplay of nodes and edges within the graph structure, efficiently encapsulating the planning knowledge of the workspace while consciously preserving object connectivity. To operationalize this approach, we employ a GNN to navigate the constructed graph, thereby ensuring that object dependencies are seamlessly integrated into the learning process when imitating motions gleaned from an oracle planner. Through rigorous comparative evaluations, we have pitted the KG-Planner against widely-used and cutting-edge planners, such as RRT, RRT*, and BFMT*. Notably, our KG-Planner exhibits promising performance in terms of both manipulator trajectory optimality and computational efficiency. The extensive experimental and comparative analyses unequivocally demonstrate that our proposed method excels in planning collision-free motions within static as well as dynamic environments, effectively merging performance and efficiency.

\ifCLASSOPTIONcaptionsoff
  \newpage
\fi

\vspace{-5pt}

\bibliographystyle{IEEEtran}

\bibliography{ref}{}

\end{document}